\documentclass[a4paper,twoside]{article}

\usepackage{amsmath}
\usepackage{amssymb}
\usepackage{amstext}
\usepackage{amsthm}
\usepackage{graphicx}
\usepackage{booktabs}
\usepackage{siunitx}
\usepackage[T1]{fontenc}
\usepackage[utf8]{inputenc}
\usepackage[table]{xcolor}
\usepackage{xspace}
\usepackage{xcolor}
\usepackage{url}
\usepackage{pifont}
\usepackage{tikz}
\usepackage{pgfplots}
\usepackage{enumitem}
\usepackage{balance}
\usepackage{soul}
\usepackage{hyperref}
\usepgfplotslibrary{fillbetween}
\pgfplotsset{compat=1.18}
\hypersetup{
	colorlinks=true,
	urlcolor=black,
	linkcolor=blue!75!black,
	citecolor=red!45!black,
} 
\usepackage{rotating} 
\usepackage{colortbl}
\newif\ifDEBUG
\DEBUGtrue

\ifDEBUG
    \newcommand{\gkt}[1]{{\small{\textcolor{purple}{\textit{\textbf{George}: #1}}} }}
    \newcommand{\laufer}[1]{{\small{\textcolor{red}{\textit{\textbf{Konstantin}: #1}}} }}
    \newcommand{\mohammed}[1]{{\small{\textcolor{blue}{\textit{\textbf{Mohammed}: #1}}} }}
    \newcommand{\taining}[1]{{\small{\textcolor{teal}{\textit{\textbf{Taining}: #1}}} }}
    \newcommand{\arslan}[1]{{\small{\textcolor{orange}{\textit{\textbf{Arslan}: #1}}} }}
    \newcommand{\brian}[1]{{\small{\textcolor{brown}{\textit{\textbf{Brian}: #1}}} }}

    \newcommand{\eric}[1]{{\small{\textcolor{brown}{\textit{\textbf{Eric}: #1}}} }}
    \newcommand{\kushboo}[1]{{\small{\textcolor{brown}{\textit{\textbf{Kushboo}: #1}}} }}

\else
    \newcommand{\gkt}[1]{}
    \newcommand{\laufer}[1]{}
    \newcommand{\mohammed}[1]{}
    \newcommand{\taining}[1]{}
    \newcommand{\arslan}[1]{}
    \newcommand{\brian}[1]{}
    \newcommand{\eric}[1]{}
    \newcommand{\kushboo}[1]{}
\fi

\usepackage{tcolorbox}
\usepackage{apalike}
\usepackage{subcaption}
\usepackage{adjustbox}
\usepackage{multirow}

\usepackage{algorithm2e}

\usepackage{SCITEPRESS}

\tcbset{
    boxsep=1mm,
    colback=gray!5,
    colframe=black,
    fonttitle=\bfseries,
    arc=1mm,
    left=0mm,
    right=0mm,
    top=0mm,
    bottom=-1mm,
    boxrule=0.2pt,
    valign=center,
}

\newcommand{\BfPara}[1]{\noindent\textbf{#1.}\xspace}

\newcommand{\eg}{{\em e.g.,}\xspace}
\newcommand{\ie}{{\em i.e.,}\xspace}

\newcommand{\tla}{{TLA$^{+}$}\xspace}

\newcommand*\cib[1]{\tikz[baseline=(char.base)]{ \node[shape=circle,fill=black,text=white,draw,inner sep=0.3pt] (char) {#1};}}

\newcommand{\UlPara}[1]{\noindent\textbf{\ul{#1:}}\xspace}

\setul{0.8px}{0.5px}

\begin{document}

\title{Can LLMs Write Correct \tla Specifications?\\ Evaluating  Natural-Language-to-\tla Generation}

\author{
\authorname{Arslan Bisharat,
Brian Ortiz,
Eric Spencer,
Khushboo Bhadauria,
TaiNing Wang,
George K.\ Thiruvathukal,
Konstantin Läufer,
Mohammed Abuhamad}
\affiliation{Department of Computer Science, Loyola University Chicago, Chicago, IL 60660, USA}
\email{\{marslan,bortiz4,espencer2,kbhadauria,twang12,gthiruvathukal,klaufer,mabuhamad\}@luc.edu}
}


\abstract{
\tla has supported industrial verification at companies such as Amazon and Microsoft, yet writing correct \tla specifications from natural language still requires time and expertise, which limits adoption. LLMs show promise, but no prior study measures whether they produce semantically correct \tla specifications from natural language.
This paper presents the first systematic evaluation of LLM-based \tla specification synthesis from natural language. 
Our study evaluates 30 LLMs across eight families on a curated dataset of 205 \tla specifications: 25 open-weight models across four prompting strategies (2,600 runs) and 5 proprietary models under few-shot prompting (130 runs), all validated by the SANY parser and TLC model checker. 
LLMs achieve up to 26.6\% syntactic correctness but only 8.6\% semantic correctness, with successes exclusive to progressive prompting. 
Results show that model size does not predict quality, \eg DeepSeek r1:8b outperforms its 70B variant across all strategies, which suggests the importance of reasoning alignment for formal languages. Code-specialized models consistently underperform due to negative transfer from mainstream language training. 
We identify five recurring hallucination categories, all traceable to specific training data biases. 
These results suggest that current LLMs do not generate reliable \tla specifications without expert oversight.
We release the evaluation framework, code, and dataset to support reproducibility and future research.} 
\onecolumn 
\maketitle 
\normalsize 
\setcounter{footnote}{0} 
\vfill

\section{Introduction} \label{sec:introduction}

Formal specification is important to build reliable distributed and concurrent 
systems~\cite{10.5555/579617}. 
Writing correct formal system specifications from natural language remains difficult and time-consuming. Large language models (LLMs) have made progress in code generation, but their ability to produce 
semantically correct formal specifications from natural language has not been 
systematically evaluated.
This work addresses \tla specification  (synthesis) from natural language with 
LLMs. 
\tla is widely adopted in industry, \eg Amazon Web Services has used it since 
2011 to verify DynamoDB, S3, and EBS~\cite{Newcombe2015}, and Microsoft Azure applied 
it to Cosmos DB~\cite{cirstea_validating_2024}. \tla combines temporal logic, 
first-order logic, and set theory, which makes it technically difficult for LLMs to 
generate correct specifications.
\tla is also a specialized formal language with a public corpus of only a few hundred modules, compared with millions of C, Python, and Java samples in standard LLM pre-training data. This data imbalance makes \tla syntax and semantics hard for LLMs to learn and leaves natural-language-to-\tla generation under-studied.

To address this gap, we curate a benchmark of 205 \tla specifications from the \tla Foundation~\cite{lamport_tla_nodate}, each paired with natural-language comments and TLC configurations, with train, validation, and test splits. We then  introduce a reproducible evaluation framework for LLM-based natural-language-to-\tla generation.
We benchmark 30 models from eight families, \ie DeepSeek, LLaMA, Qwen, QwQ, GPT-OSS, code-specialized (\eg CodeLLaMA and Granite), instruction-tuned (\eg Mistral, Phi, Gemma, and Starling-LM), and proprietary APIs (\eg OpenAI GPT and Anthropic Claude) on the curated dataset.
The core evaluation uses 25 
open-weight models across four prompt strategies (2,600 runs). We also evaluate five 
proprietary models with few-shot ($k=3$) only (130 runs). All outputs from all runs are validated by 
the SANY parser and TLC checker.

\emph{Key Findings.}
(1) LLMs achieve up to 26.6\% syntactic correctness on \tla (SANY) but only 8.6\% 
semantic correctness (TLC). Semantic success is exclusive to progressive prompts. 
(2) Model size does not predict quality. The 8B-parameter DeepSeek~r1 
outperforms its 70B sibling across all strategies. Reasoning-trace fine-tuning 
is a stronger predictor of quality than model size. 
(3) Code-specialized models consistently underperform general-purpose ones due to negative 
transfer (due to programming language bias) from mainstream programming logic and language training. 
(4) We identify five recurring hallucination categories: Unicode operator substitution, cross-language syntax injection, reasoning and formatting leakage, generation length miscalibration, and structural errors. These errors map to biases in current open-source training data (\ie code, formal math, and reasoning samples) 
and point to two mitigation insights, \eg obtaining high-quality datasets for specifications and grammar-constrained generation.

\BfPara{Contributions} Our contributions are: 
    \ding{182} We curate a dataset for natural-language-to-\tla specification synthesis, consisting of 205 \tla specifications paired with natural language comments and TLC configurations.    \ding{183} We provide a systematic evaluation of LLM-based \tla specification generation/synthesis from natural language, testing 30 models from eight families across four prompting strategies with  syntactic and semantic validation. \ding{184} We establish quantified baselines for \tla synthesis and highlight the gap between current LLM capabilities and reliable specification synthesis. We provide a taxonomy of five systematic hallucination categories in \tla generation/synthesis and offer directions for improvement. We release code, dataset, models, and results to support future work.

\section{Background \& Related Work}\label{sec:background}

\BfPara{\tla and Formal Specification} Distributed and concurrent systems are difficult to design and verify. Small errors in the system's logic can create bugs that appear only under rare timing conditions, so standard tests often cannot find them. Formal verification solves this problem by the application of mathematical reasoning to prove that a system behaves correctly in every possible scenario. \tla (Temporal Logic of Actions) \cite{10.5555/579617} was created to describe concurrent and distributed systems behavior through mathematical formulas that combine temporal logic, first-order logic and set theory.

A \tla specification has several key components. The initial predicate \textit{Init} defines all valid start states and the next-state relation \textit{Next} describes how the system moves from one state to another through atomic actions. The expression \textit{Init} $\land$ $\square$ [\textit{Next}]$_v$ defines the core behavior of the system. The symbol $\square$ represents ``always.'' System properties are expressed as invariants (safety properties that hold in every reachable state) and liveness properties (behaviors that must eventually occur). The \tla toolchain provides two verification methods: the TLC model checker examines every reachable state in a finite version of a specification~\cite{10.5555/579617}, while TLAPS allows machine-checked deductive reasoning for infinite-state specifications~\cite{cirstea_validating_2024}. 
Although \tla is effective at the detection of design flaws, the creation of correct specifications requires specialized expertise. Engineers must translate informal descriptions with ambiguities and hidden assumptions into precise mathematical expressions that resolve questions about failure modes, concurrency and consistency. Small mistakes such as an incorrect quantifier bound or a missing fairness condition can make a specification compile but fail to capture intended behavior. This expertise barrier limits \tla adoption.

\BfPara{LLM-based Generation} LLMs trained on large collections of source code can generate code from text. Models such as GPT-4, Gemini, Claude and Code Llama can solve standard benchmarks like HumanEval  \cite{austin2021program}. But LLMs still make many mistakes: they sometimes create APIs or library functions that do not exist, produce code that does not follow the rules in the specification and struggle with problems that require several steps of reasoning \cite{ye_how_2023}. The generation of formal specifications is even harder than code creation. A normal program can be checked by execution, but a formal specification must follow strict mathematical rules. A \tla specification may be written correctly with proper operators and syntax but still describe the wrong system behavior by leaving out a fairness rule, defining a condition that is too weak, or leaving important variables undefined. These problems cannot be found by parsers or type checkers and show up only during formal verification or detailed expert review. LLM hallucination, where a model confidently produces incorrect output, has been classified for software tasks into task requirement conflicts, factual knowledge conflicts and project context conflicts~\cite{zhang2025llmhallucinations}. Package hallucination creates supply chain risks~\cite{spracklen2025packagehallucinations}. For formal languages like \tla, hallucinated errors may only surface during verification.

\BfPara{GenAI for \tla} The \tla Foundation launched the GenAI-accelerated \tla Challenge in 2025 to investigate generative AI applications for formal methods \cite{TLAFoundation2025}. Specula, the winner, generates \tla specifications from code that already exists \cite{specula2025}. The system converts source code into intermediate \tla representations, fixes errors through Retrieval-Augmented Generation and iteratively corrects runtime errors identified by TLC. The creation of the Raft specification took about 1.5 hours of manual effort. The second-place entry tested whether LLMs can be guided to produce valid \tla syntax through token-level rules such as Greibach Backus-Naur Form and the Guidance framework \cite{helwer2025}. This method focuses primarily on the production of syntactically correct specifications while semantic correctness is not addressed. Gregory Terzian studied the opposite approach: the use of completed \tla specifications to guide LLMs in the generation of Rust code \cite{terzian2025}. SysMoBench~\cite{cheng2025sysmobenchevaluatingaiformally} is the first benchmark for LLMs on complex distributed systems in \tla, but it compares generated specifications to source code implementations. Our work evaluates generation from natural language only. Previous specification synthesis work focused on invariant discovery from program traces: DistAI infers inductive invariants for distributed protocols~\cite{273755} and SpecGen uses LLMs to generate function-level Java specifications through iterative verification feedback~\cite{SpecGen}. Neither generates system-level specifications like \tla.

\BfPara{Gap in Prior Work and Our Contribution} Existing GenAI research on \tla focuses on the creation of specifications from code, the enforcement of syntax through token constraints and the use of specifications to guide code generation. No systematic study has tested whether LLMs can produce full and semantically correct \tla specifications directly from natural language, what error patterns emerge across model architectures, or how prompting strategies affect generation quality. 
This work provides the first systematic evaluation of 30 LLMs across four prompting strategies to measure both syntactic and semantic correctness of \tla specifications generated from natural language. We use the \tla Examples repository, which contains both formal specifications and natural language comments, to test which model architectures and prompting approaches produce correct specifications and what types of errors and hallucinations occur.

\section{Methods} \label{sec:methods}

\BfPara{Research Questions} 
This research addresses four questions. 
\textbf{RQ1:} \emph{Can LLMs generate \tla specifications that are both 
syntactically correct and semantically faithful?} \textbf{RQ2:} \emph{What error 
types do different LLM backends produce?}
\textbf{RQ3:} \emph{How do quality metrics 
correlate with model architecture, size, and fine-tuning?} 
\textbf{RQ4:} \emph{What 
failure modes and hallucinations recur in LLM-generated specifications?} Formal 
validation (SANY and TLC) answers RQ1, RQ2, and RQ4. Textual similarity 
metrics answer RQ3.
No public dataset supports systematic evaluation of LLM-generated \tla 
specifications across these four dimensions. We curate a benchmark of 205 
specifications to enable this evaluation.

\begin{table}[t]
\centering
\caption{Summary of data availability across 205 TLA+ specifications in our dataset.}
\label{tab:dataset-summary}
\resizebox{0.4\textwidth}{!}{
\begin{tabular}{lcc}
\toprule
\textbf{Category} & \textbf{Count} & \textbf{Percentage} \\
\midrule
Both CFG \& Comments & 92 & 44.9\% \\
CFG only & 14 & 6.8\% \\
Comments only & 95 & 46.3\% \\
Neither & 4 & 2.0\% \\
\midrule
Has CFG & 106 & 51.7\% \\
Has Comments & 187 & 91.2\% \\
\bottomrule
\end{tabular}}
\end{table}

\BfPara{Dataset} \label{sec:dataset} We use the \tla Examples repository 
maintained by the \tla Foundation~\cite{lamport_tla_nodate}, the primary 
source of community-written formal specifications. The repository contains 205 
\tla specifications across 98 projects: distributed consensus protocols 
(Paxos, Raft, Byzantine agreement), concurrency problems (dining philosophers, 
readers-writers), and algorithmic puzzles (Towers of Hanoi, N-queens). Each 
specification includes a \texttt{.tla} file, extracted natural language comments 
used as LLM input, and an optional TLC configuration file. 
We split the dataset into training (143 specifications, 70\%), validation (31 specifications, 15\%), and test (31 specifications, 15\%) sets stratified by project. 
The training set provides few-shot examples. 
The validation set is used for model hyperparameters and prompt tuning.
The test set originally 
contains 31 specifications. Five were excluded because they lacked either natural 
language comments or a TLC configuration file, both required for evaluation. 
This yields 26 usable test specifications. Table ~\ref{tab:dataset-summary} 
shows the distribution across splits.

\begin{table}[t]
\centering
\caption{Token usage by model family. R-Tok: reasoning tokens; T-Tok: total tokens; R\%: reasoning percentage; Ctx: context window size.}
\label{tab:token_breakdown_family}
\resizebox{0.4\textwidth}{!}{
\begin{tabular}{@{}lp{2cm}rrrr@{}}
\toprule
\textbf{Family} & \textbf{Model} & \textbf{R-Tok} & \textbf{T-Tok} & \textbf{R\%} & \textbf{Ctx} \\
\midrule
\multicolumn{6}{c}{\textit{\textbf{Proprietary Models}}} \\
\midrule
GPT & gpt-5 & 3.8K & 24K & 15.8 & 128K \\
 & gpt-4o & 338 & 19K & 1.8 &  128K\\
\cmidrule(lr){1-6}
Claude & sonnet-4-5 & 338 & 28K & 1.2 & 200K \\
 & haiku-4-5 & 338 & 17K & 2.0 &   200K\\
 & opus-4-1 & 338 & 31K & 1.1 &  200K\\
\midrule
\multicolumn{6}{c}{\textit{\textbf{Open-Weight Models}}} \\
\midrule
DeepSeek & r1:70b & 364 & 12K & 3.0 & 131K \\
 & r1:32b & 338 & 12K & 2.8 &  131K\\
 & r1:14b & 338 & 9.5K & 3.5 &  131K\\
 & r1:8b & 390 & 6.7K & 5.8 &  131K\\
 & r1:7b & 364 & 25K & 1.4 &  131K\\
 & r1:1.5b & 390 & 1141.7K& 0.03 &  131K\\
 & coder:6.7b & 390 & 754K & 0.05 &  4K\\
 & coder:latest & 286 & 289K & 0.1 &  128K\\
\cmidrule(lr){1-6}
LLaMA & 3.3:70b & 338 & 20K & 1.7 & 128K \\
 & 3.3:latest & 338 & 20K & 1.7 &  128K\\
 & 3.1:8b & 416 & 17K & 2.4 &  128K\\
 & 3:latest & 338 & 16K & 2.1 &  8K\\
 & 3.2:1b & 390 & 771K & 0.05 &  131K\\
\cmidrule(lr){1-6}
Qwen & 3:235b & 520& 18K&  2.9&  32K\\
 & 3:30b & 520& 17K& 3.1&  32K\\
 & 3:4b & 390 & 4.1K & 9.6 &  \\
 \cmidrule(lr){1-6}

QwQ & latest & 390 & 110K & 0.4 & 32K \\
Phi & 4-mini:latest & 416 & 1.3K & 31.7 & 128K \\
Granite & 3.3:latest & 390 & 42K & 0.9 & 131K \\
GPT-OSS & 20b & 520& 20K& 2.6& 128K \\
 & 120b & 390 & 19K & 2.1 &  128K\\
CodeLLaMA & latest & 416 & 56K & 0.8 & 16K \\
Mistral & latest & 390 & 61K & 0.6 & 32K \\
Gemma & 2b & 520& 29K& 1.8& 8K \\
Starling-LM & latest & 364 & 38K & 1.0 & 8K \\
\bottomrule
\end{tabular}}
\end{table}

\BfPara{LLM Models}  We test 30 LLMs from eight model 
families (Table~\ref{tab:token_breakdown_family}). The core evaluation uses 25 open-weight models across all four 
strategies via the Ollama runtime~\cite{ollama2024}: 
\ding{182}~\textit{DeepSeek} (R1 series: 1.5B--70B; 
Coder)~\cite{Guo_2025}; 
\ding{183}~\textit{LLaMA} (3.x: 
1B--70B)~\cite{grattafiori2024llama3herdmodels}; 
\ding{184}~\textit{Code-specialized} (CodeLLaMA, DeepSeek-Coder, Granite); 
\ding{185}~\textit{Reasoning-focused} (QwQ, 
Qwen3~\cite{yang2025qwen3technicalreport}); 
\ding{186}~\textit{Instruction-tuned} (Mistral~\cite{mistral2023}, Starling-LM, 
Phi-4~\cite{abouelenin2025phi}, 
Gemma~2b~\cite{gemmateam2024gemma}). We also test five proprietary models with 
few-shot ($k=3$) prompting: GPT-5~\cite{singh2025openaigpt5card}, 
GPT-4o~\cite{openai2024gpt4technicalreport}, Claude 
Opus~4.1~\cite{anthropic2023claude}, Sonnet~4.5~\cite{sonnet45}, and Haiku~4.5. We focus the 
multi-strategy analysis on open-weight models for two reasons. First, 
open-weight models support full reproducibility. Second, formal specifications 
often describe sensitive system designs better suited to local generation.

\BfPara{Prompting Strategies} 
We test four prompting 
strategies adapted from common LLM code-generation settings for \tla specification generation/synthesis. Each approach tests different LLM capabilities.

\noindent\cib{1} \UlPara{Few-Shot Prompting} This strategy tests few-shot in-context learning 
with $k=3$ examples from the training set. Each example pairs natural language 
comments with the corresponding complete \tla specification. The prompt 
consists of a system instruction, the few-shot examples, and the target comments. 
The LLM generates the full specification in one step.

\noindent\cib{2} \UlPara{Progressive Prompting} This strategy also uses $k=3$ few-shot examples 
but adds instructional messages that guide the model through the specification 
structure. Each message focuses on a different aspect: module declaration, state 
variables, operators, and temporal properties. This structure prepares the model 
before it receives the target comments.

\noindent\cib{3} \UlPara{Fill-in-Middle} This code completion strategy splits the ground-truth 
specification into three parts: prefix (first 30\%), middle (40\%), and suffix 
(last 30\%). The prompt provides the prefix and suffix with a \texttt{<FILL>} 
marker. The LLM generates the missing middle portion. No few-shot examples are 
used. The surrounding context provides sufficient syntactic and semantic 
information for the model.

\noindent\cib{4} \UlPara{Half Completion} This strategy provides the first 50\% of the 
specification as a prefix along with comments and configuration. The LLM 
generates the remaining 50\%. No few-shot examples are used. The prefix 
provides sufficient syntactic and semantic context for the model.

\BfPara{Evaluation Metrics}
We use two metric groups with different roles. Formal validation measures correctness of the generated specification as an executable formal artifact.  Textual similarity measures closeness to the reference sequence. 

\UlPara{Formal Validation} Each specification passes through two stages. 
\ding{182}~SANY checks syntax with a 30-second timeout and passes if it reports 
``Parsing completed.'' \ding{183}~Specifications that pass SANY are submitted 
to TLC with their configuration file for model checking.

\UlPara{Textual Similarity} For completion strategies (HC and FIM), a 
well-defined target region exists for direct comparison. We compute 
BLEU~\cite{papineni-etal-2002-bleu}, ROUGE-L~\cite{lin-2004-rouge}, edit 
distance, exact match, and line accuracy against the ground-truth segment.

\begin{table}[t]
\centering
\caption{Cross-method comparison of SANY, TLC and textual similarity results across four prompting strategies (25 open-weight models $\times$ 26 specifications = 650 runs per strategy). Only Progressive achieves TLC passes (8.6\%). Per-model breakdowns appear in Table~\ref{tab:all-methods-results}; per-specification breakdowns in Table~\ref{tab:spec-summary}. $^\dagger$TLC pass rates of 0\% reflect failures (\ie State Directory Conflicts and Module Not Found errors).}
\label{tab:cross-method-summary}
\resizebox{0.48\textwidth}{!}{
\begin{tabular}{@{}l |c |c |c |c@{}}
\toprule
 & \textbf{FS} & \textbf{HC} & \textbf{Prog.} & \textbf{FIM} \\
\midrule
\multicolumn{5}{c}{\textit{\textbf{Syntax Validation (SANY)}}} \\
\midrule
Passed / Total          & 173 / 650   & 80 / 650     & 162 / 650    & 92 / 650 \\
Pass Rate (\%)          & 26.6        & 12.2         & 24.9         & 14.2 \\
Best Model (\%)         & qwen3:30b (92.3) & qwq (34.6)   & qwen3:235b (80.8) & qwq (38.5) \\
Specs at 0\%            & 0 / 26      & 6 / 26       & 0 / 26       & 9 / 26 \\
\midrule
\multicolumn{5}{c}{\textit{\textbf{Model Checking (TLC)}}} \\
\midrule
Passed / Total          & 0 / 650     & 0 / 650        & 56 / 650       & 0 / 650 \\
Pass Rate (\%)          & 0.0 & 0.0 & 8.6            & 0.0 \\
Best Model (\%)         & N/A         & N/A            & r1:8b (53.8)   & N/A \\
Specs at 0\%            & 26 / 26     & 26 / 26        & 1 / 26         & 26 / 26 \\
\midrule
\multicolumn{5}{c}{\textit{\textbf{Similarity to Ground Truth}}} \\
\midrule
Avg BLEU                & N/A         & $0.062_{\pm 0.11}$       & N/A      &  $0.077_{\pm 0.13}$\\
Avg ROUGE-L             & N/A         & $0.189_{\pm 0.19}$       & N/A      &  $0.213_{\pm 0.20}$\\
Avg Line Accuracy (\%)  & N/A         & $10.90_{\pm 11.8  }$     & N/A      & $12.27_{\pm 13.9}$ \\
Exact Matches           & N/A         & 0 / 650      & N/A      & 1 / 650 \\
\midrule
\multicolumn{5}{l}{\textbf{FS}: Few Shot, \textbf{HC}: Half Completion, \textbf{Prog.}: Progressive, \textbf{FIM}: Fill-in-Middle}\\
\bottomrule
\end{tabular}
}
\end{table}

\section{Results} \label{sec:results}

Table~\ref{tab:cross-method-summary} summarizes results across all four 
prompting strategies. The headline finding is a persistent syntax--semantics 
gap: \ie even the best strategy reaches only 26.6\% SANY syntax validity. Only 
Progressive prompting produces any model-checking successes (8.6\%). Within 
those successes, a single 8B reasoning model accounts for a quarter of all 
passing runs and outperforms every larger model tested.

\begin{table*}[t]
\centering
\caption{TLC model checking infrastructure failure analysis across prompting strategies. Progressive shows 122 TLC failures plus 455 parsing errors (577 total failures, 56 passes, 8.6\% pass rate). FS, HC and FIM each had 650 TLC failures (0\% pass rate). The table decomposes TLC failures into infrastructure artifacts (Module Not Found) and genuine semantic failures (Config File Exception: 18 FS, 5 HC, 110 Prog., 5 FIM). 
}
\label{tab:tlc-infrastructure}

\resizebox{0.85\textwidth}{!}{

\begin{tabular}{@{}l l r r r r l@{}}
\toprule

\textbf{Category} & \textbf{Type} & \textbf{FS} & \textbf{HC}& \textbf{Prog.}& \textbf{FIM} & \textbf{Description} \\
\midrule

Module Not Found     & Infra. & 159 (24.5\%) & 159 (24.5\%) & 12 (9.8\%) & 156 (24.0\%) & Unresolved \texttt{EXTENDS} module paths \\
Config File Exception & \textbf{Genuine} & 18 (2.8\%) & 5 (0.8\%) & 110 (90.2\%) & 5 (0.8\%) & Invalid \texttt{.cfg} for TLC \\
\midrule
\textbf{Total} & & \textbf{650} & \textbf{650} & \textbf{122} & \textbf{650} & \\
Infrastructure \%    & & 97.2\% & 99.2\% & 9.8\% & 99.2\% & \\
\bottomrule
\end{tabular}
}
\end{table*}\

\begin{table}[t]
\centering
\caption{Per-model SANY (S) and TLC (T) pass counts across all four prompting strategies (25 open-weight models, 26 specifications each). Detailed per-model per-specification breakdowns are provided in supplementary materials.}
\label{tab:all-methods-results}

\resizebox{0.48\textwidth}{!}{
\begin{tabular}{@{}l|p{1.75cm}|c@{}c|c@{}c|c@{}c|c@{}c@{}}
\toprule
 & & \multicolumn{2}{c|}{\textbf{FS}} & \multicolumn{2}{c|}{\textbf{HC}} & \multicolumn{2}{c|}{\textbf{Prog.}} & \multicolumn{2}{c}{\textbf{FIM}} \\
\cmidrule(lr){3-4} \cmidrule(lr){5-6} \cmidrule(lr){7-8} \cmidrule(lr){9-10}
\textbf{Family} & \textbf{Model} & \textbf{S} & \textbf{T} & \textbf{S} & \textbf{T} & \textbf{S} & \textbf{T} & \textbf{S} & \textbf{T} \\
\midrule

 & r1:70b & 0 & 0 & 3 & 0 & 3 & 0 & 3 & 0 \\

 & r1:32b & 6 & 0 & 2 & 0 & 8 & 1 & 6 & 0 \\

 & r1:14b & 1 & 0 & 2 & 0 & 2 & 1 & 2 & 0 \\

 DeepSeek & r1:8b & 22 & 0 & 1 & 0 & 15 & 14 & 3 & 0 \\

 & r1:7b & 1 & 0 & 4 & 0 & 6 & 1 & 3 & 0 \\

 & r1:1.5b & 1 & 0 & 2 & 0 & 8 & 3 & 2 & 0 \\

 & coder:6.7b & 1 & 0 & 1 & 0 & 0 & 0 & 2 & 0 \\

 & coder:latest & 0 & 0 & 5 & 0 & 3 & 1 & 5 & 0 \\
\midrule
 & 3.3:70b & 11 & 0 & 2 & 0 & 18 & 0 & 6 & 0 \\

 & 3.3:latest & 11 & 0 & 1 & 0 & 18 & 0 & 5 & 0 \\

LLaMA & 3.1:8b & 6 & 0 & 3 & 0 & 3 & 0 & 4 & 0 \\

 & 3:latest & 3 & 0 & 3 & 0 & 2 & 0 & 5 & 0 \\

 & 3.2:1b & 2 & 0 & 2 & 0 & 12 & 0 & 4 & 0 \\
\midrule
 & 3:235b & 22 & 0 & 3 & 0 & 21 & 12 & 5 & 0 \\

 Qwen & 3:30b & 23 & 0 & 3 & 0 & 11 & 10 & 5 & 0 \\

 & 3:4b & 5 & 0 & 3 & 0 & 1 & 1 & 2 & 0 \\
\midrule
QwQ & latest & 3 & 0 & 9 & 0 & 2 & 0 & 10 & 0 \\
\midrule
Phi & 4-mini:latest & 5 & 0 & 4 & 0 & 4 & 4 & 1 & 0 \\
\midrule
Granite & 3.3:latest & 3 & 0 & 5 & 0 & 2 & 0 & 2 & 0 \\
\midrule
GPT-OSS & 120b & 17 & 0 & 4 & 0 & 6 & 1 & 3 & 0 \\

 & 20b & 18 & 0 & 3 & 0 & 9 & 6 & 5 & 0 \\
\midrule
CodeLLaMA & latest & 3 & 0 & 5 & 0 & 2 & 0 & 3 & 0 \\
\midrule
Mistral & latest & 3 & 0 & 2 & 0 & 3 & 0 & 3 & 0 \\
\midrule
Gemma & 2b & 6 & 0 & 4 & 0 & 2 & 1 & 2 & 0 \\
\midrule
Starling-LM & latest & 0 & 0 & 2 & 0 & 1 & 0 & 1 & 0 \\
\midrule
\multicolumn{2}{l|}{\textbf{Average}} 
& \textbf{6.92} 
& \textbf{0.00} 
& \textbf{3.17} 
& \textbf{0.00} 
& \textbf{6.38} 
& \textbf{2.08} 
& \textbf{3.71} 
& \textbf{0.00} \\
\midrule
\multicolumn{10}{l}{\textbf{FS}: Few Shot, \textbf{HC}: Half Completion, \textbf{Prog.}: Progressive, \textbf{FIM}: Fill-in-Middle}\\
\bottomrule
\end{tabular}
}

\end{table}

\subsection{RQ1: Synthesis Correctness}

\UlPara{Syntax Validation} FS achieves the highest SANY pass rate at 26.6\% 
(173/650) followed by Prog.\ at 24.9\% (162/650). The two completion 
strategies lag behind: FIM passes 14.2\% (92/650) and HC passes 12.2\% 
(80/650).

\UlPara{Model Checking} The clearest result is the near-total failure of 
semantic validation. Three of the four strategies (FS, HC, and FIM) achieve 
0\% TLC pass rates across all 650 runs each. Prog.\ is the only exception 
with 56 of 650 runs (8.6\%) passing TLC model checking 
(Table~\ref{tab:tlc-infrastructure} details two categories of TLC failure:
\textit{infrastructure failures}, \ie environment-level issues such as unresolved
\texttt{EXTENDS} module paths that prevent TLC from loading the specification,
and \textit{genuine semantic failures} caused by invalid model-checking configuration
in the LLM output). The 56 passes are 
concentrated in just three model families. DeepSeek r1:8b is the strongest 
performer with 14 out of 26 TLC passes (53.8\%) under Prog., a result that 
accounts for 25\% of all successful runs. Qwen3:235b and Qwen3:30b follow 
with 12 and 10 TLC passes, respectively, together contributing another 39\% 
of all Prog.\ successes. GPT-OSS-20b achieves 6 and Phi-4-mini 4 TLC passes; 
every other model achieves 3 or fewer. The Qwen family's TLC success under 
Prog.\ contrasts sharply with its 0 TLC passes under FS despite its highest 
FS syntax rates. We examine this pattern in RQ3.
\UlPara{Best Models per Strategy} Table~\ref{tab:all-methods-results} 
shows two contrasting performance profiles. For full-generation strategies 
(FS and Prog.), large general-purpose models dominate: Qwen3:30b (23/26) and 
Qwen3:235b (22/26) lead FS SANY, while Qwen3:235b (21/26) and both 
LLaMA~3.3 variants (18/26) lead Prog.\ SANY. For completion strategies, the 
picture inverts: QwQ achieves only 3/26 on FS yet leads both HC (9/26) and 
FIM (10/26). Chain-of-thought reasoning appears to help context-sensitive 
completion, but hurts open-ended generation.

\UlPara{Proprietary Models (FS $k=3$ Only)} In a companion experiment, we 
test five proprietary models with FS ($k=3$) as the sole strategy. 
The detailed results are shown in 
Table~\ref{tab:proprietary-summary}. Among these, GPT-5 is the strongest 
overall: it passes SANY on all 26 specifications (100\%) and achieves 7/26 
TLC passes (26.9\%), the highest TLC rate for any model under FS. Claude 
Sonnet~4.5 and Haiku~4.5 pass SANY at 92\% (24/26) and 81\% (21/26) 
respectively and each achieves 3/26 TLC passes (11.5\%). Claude Opus~4.1 
passes 23/26 SANY (88.5\%) with 1 TLC pass. GPT-4o passes 20/26 SANY 
(77\%) with 1 TLC pass. These proprietary models were not included in the 
multi-strategy analysis due to reproducibility constraints and because formal 
specifications often describe sensitive system designs. Even the best frontier 
model (GPT-5) achieves 26.9\% TLC success under FS, comparable to the 
open-weight FS ceiling. Syntax--semantics gap persists at frontier scale 
under the same prompting conditions. 

\begin{table}[t]
\centering
\caption{SANY and TLC performance for five proprietary LLMs under Full-Generation (FS) prompting. GPT-5 achieves the highest TLC rate at 26.9\%, confirming that the syntax--semantics gap persists even at frontier model scale.}
\label{tab:proprietary-summary}
\resizebox{0.4\textwidth}{!}{
\begin{tabular}{@{}l|cc|cc@{}}
\toprule
\textbf{Model} & \multicolumn{2}{c|}{\textbf{SANY}} & \multicolumn{2}{c}{\textbf{TLC}} \\
\cmidrule(lr){2-3} \cmidrule(lr){4-5}
 & \textbf{Pass} & \textbf{\%} & \textbf{Pass} & \textbf{\%} \\
\midrule
GPT-5 & 26/26 & 100.0 & 7/26 & 26.9 \\
Claude Sonnet 4.5 & 24/26 & 92.3 & 3/26 & 11.5 \\
Claude Haiku 4.5 & 21/26 & 80.8 & 3/26 & 11.5 \\
Claude Opus 4.1 & 23/26 & 88.5 & 1/26 & 3.8 \\
GPT-4o & 20/26 & 76.9 & 1/26 & 3.8 \\
\bottomrule
\end{tabular}}
\end{table}

\UlPara{Textual Similarity} For the completion strategies where ground-truth 
comparison is possible, similarity to the reference is low as shown in Tables~\ref{tab:half-completion-summary} and \ref{tab:fill-in-middle-summary}. HC achieves a 
mean BLEU of 0.062 ($\pm$0.11) and ROUGE-L of 0.189 ($\pm$0.19) while FIM 
scores slightly higher with BLEU of 0.077 ($\pm$0.13) and ROUGE-L of 0.213 
($\pm$0.20). Line accuracy averages 10.90\% for HC and 12.27\% for FIM; 
exact matches are nearly absent (0/650 for HC, 1/650 for FIM). These low 
scores follow directly from the length miscalibration documented in RQ4: 
models that generate 3--9x the expected output length will diverge 
from ground-truth n-gram patterns regardless of semantic quality. 

\begin{tcolorbox}[title=\textbf{RQ1 Summary}]
LLMs produce syntactically valid \tla in up to 26.6\% of attempts, but semantic 
correctness is rare (8.6\% TLC under Prog.). 
Only Prog.\ achieves any TLC 
passes, led by DeepSeek r1:8b (14/26) and Qwen3 models (10--12 TLC each). 
Completion strategies show low 
textual similarity (BLEU $<$ 0.08 and ROUGE-L $<$ 0.22). 
\end{tcolorbox}

\subsection{RQ2: Synthesis Errors}

\UlPara{Dominant Error Categories} Table~\ref{tab:error-taxonomy} (Panel-A) 
shows the distribution of SANY errors across the four strategies. The most 
common error category is \textit{Parse: Bad Module Body}, which accounts for 
roughly half of all failures across every strategy (47.8\% FS, 46.3\% HC, 
49.8\% Prog., 52.0\% FIM). This error arises when the parser encounters 
invalid syntax within the module body: malformed operator definitions, 
structurally broken expressions, or injected non-\tla syntax. 
This pattern remains constant across strategies, which indicates a core limitation in LLM understanding of \tla grammar rather than prompt design.

\UlPara{Strategy-Specific Error Patterns} The second most common error type 
differs by strategy class. FS and Prog.\ produce relatively more 
\textit{Unexpected Token} errors (17.2\% and 13.3\%), which arise when the 
parser encounters a token that violates the expected grammar. In contrast, HC 
and FIM produce more \textit{Parse: Other} errors (39.8\% and 34.2\%), which 
capture miscellaneous parse failures such as mismatched delimiters and 
incomplete expressions. Full-generation models must construct correct operator 
syntax from scratch and tend to fail with specific illegal tokens. Completion 
models receive valid syntactic context and fail with structural mismatches 
deeper in the file.

\UlPara{Lexical Errors} Lexical errors account for less than 10\% of 
SANY-detected failures but reveal systematic cross-language contamination. 
Unicode operator substitution (replacing \tla ASCII operators with 
\LaTeX/math equivalents) appears in 7.1\% of FS and 8.8\% of Prog.\ 
failures. Backtick errors from Markdown code-fence leakage appear in 5.9\% 
of FS failures. Semicolons, absent from \tla but ubiquitous in C, Java, 
and Python, appear as SANY first errors in 3.8\% of FS and 5.1\% of Prog.\ 
failures. A broader file scan reveals higher ambient contamination rates of 
7.5\% for FS and 12.6\% for Prog. These lexical hallucinations are discussed 
in full under RQ4.

\UlPara{Error Location} Panel~B of Table~\ref{tab:error-taxonomy} shows 
where the first error occurs in the generated file, divided into early (first 
third), middle, and late (final third) segments. FS and Prog.\ exhibit 
early-heavy error distributions: 50.3\% and 45.1\% of errors occur in the 
first third of the file. HC and FIM show the opposite pattern: errors 
concentrate in the middle and late portions (88.9\% and 85.5\% combined). 
Full-generation strategies must produce a valid \tla module header from 
scratch and fail immediately when the preamble is wrong. Completion strategies 
receive a correct prefix and only diverge further into the body.

\begin{tcolorbox}[title=\textbf{RQ2 Summary}]
``Parse: Bad Module Body'' accounts for roughly half of all failures across every 
strategy and points to a structural failure mode that future datasets should 
target. Full-generation strategies fail early with specific illegal tokens; 
completion strategies fail mid-to-late with structural mismatches. Lexical 
contamination from non-\tla languages (Unicode operators, backticks, 
semicolons) is systematic and strategy-independent. Grammar-constrained 
generation and deterministic post-processing offer concrete paths for future 
improvement.
\end{tcolorbox}

\begin{table}[t]
\centering
\caption{Per-model textual similarity metrics for Fill-in-the-Middle (FIM) prompting (mean$_{\pm}$SD across 26 specifications). Includes BLEU, ROUGE, edit distance, exact match and line accuracy.}
\label{tab:fill-in-middle-summary}
\resizebox{0.48\textwidth}{!}{
\begin{tabular}{@{}l|p{1.75cm}|c|c|c|c|c@{}}
\toprule
\textbf{Family} & \textbf{Model} & \textbf{E} & \textbf{Acc (\%)} & \textbf{Similarity} & \textbf{BLEU} & \textbf{ROUGE-L} \\
\midrule
 & r1:70b & 1 & $18.59_{\pm 21.2}$ & $0.151_{\pm 0.19}$ & $0.159_{\pm 0.21}$ & $0.348_{\pm 0.22}$ \\
 & r1:32b & 0 & $19.02_{\pm 20.0}$ & $0.195_{\pm 0.18}$ & $0.145_{\pm 0.17}$ & $0.371_{\pm 0.19}$ \\
 & r1:14b & 0 & $12.19_{\pm 14.2}$ & $0.152_{\pm 0.18}$ & $0.076_{\pm 0.13}$ & $0.270_{\pm 0.20}$ \\
DeepSeek & r1:8b & 0 & $6.04_{\pm 10.0}$ & $0.028_{\pm 0.05}$ & $0.017_{\pm 0.03}$ & $0.048_{\pm 0.07}$ \\
 & r1:7b & 0 & $8.06_{\pm 9.4}$ & $0.128_{\pm 0.12}$ & $0.053_{\pm 0.08}$ & $0.217_{\pm 0.12}$ \\
 & r1:1.5b & 0 & $7.89_{\pm 11.0}$ & $0.064_{\pm 0.09}$ & $0.027_{\pm 0.05}$ & $0.131_{\pm 0.16}$ \\
 & coder:6.7b & 0 & $13.33_{\pm 17.5}$ & $0.076_{\pm 0.11}$ & $0.044_{\pm 0.07}$ & $0.209_{\pm 0.19}$ \\
 & coder:latest & 0 & $9.55_{\pm 12.8}$ & $0.075_{\pm 0.09}$ & $0.023_{\pm 0.04}$ & $0.171_{\pm 0.15}$ \\
\midrule
 & 3.3:70b & 0 & $22.39_{\pm 17.1}$ & $0.194_{\pm 0.16}$ & $0.204_{\pm 0.19}$ & $0.377_{\pm 0.17}$ \\
 & 3.3:latest & 0 & $21.70_{\pm 18.4}$ & $0.185_{\pm 0.17}$ & $0.195_{\pm 0.19}$ & $0.367_{\pm 0.18}$ \\
LLaMA & 3.1:8b & 0 & $9.55_{\pm 11.9}$ & $0.132_{\pm 0.13}$ & $0.121_{\pm 0.15}$ & $0.278_{\pm 0.16}$ \\
 & 3:latest & 0 & $10.68_{\pm 11.3}$ & $0.142_{\pm 0.14}$ & $0.091_{\pm 0.13}$ & $0.285_{\pm 0.15}$ \\
 & 3.2:1b & 0 & $3.82_{\pm 6.0}$ & $0.075_{\pm 0.09}$ & $0.026_{\pm 0.05}$ & $0.135_{\pm 0.14}$ \\
\midrule
 & 3:235b & 0 & $9.62_{\pm 13.1}$ & $0.054_{\pm 0.09}$ & $0.054_{\pm 0.08}$ & $0.087_{\pm 0.14}$ \\
Qwen & 3:30b & 0 & $5.37_{\pm 6.8}$ & $0.033_{\pm 0.06}$ & $0.013_{\pm 0.02}$ & $0.049_{\pm 0.08}$ \\
 & 3:4b & 0 & $8.27_{\pm 11.7}$ & $0.149_{\pm 0.16}$ & $0.021_{\pm 0.04}$ & $0.191_{\pm 0.16}$ \\
\midrule
QwQ & latest & 0 & $14.66_{\pm 19.5}$ & $0.030_{\pm 0.05}$ & $0.020_{\pm 0.04}$ & $0.070_{\pm 0.08}$ \\
\midrule
Phi & 4-mini:latest & 0 & $2.20_{\pm 6.4}$ & $0.023_{\pm 0.05}$ & $0.001_{\pm 0.00}$ & $0.025_{\pm 0.07}$ \\
\midrule
Granite & 3.3:latest & 0 & $14.45_{\pm 13.5}$ & $0.130_{\pm 0.13}$ & $0.103_{\pm 0.14}$ & $0.285_{\pm 0.17}$ \\
\midrule
GPT-OSS & 120b & 0 & $24.16_{\pm 21.0}$ & $0.240_{\pm 0.24}$ & $0.212_{\pm 0.25}$ & $0.392_{\pm 0.25}$ \\
 & 20b & 0 & $8.46_{\pm 12.7}$ & $0.065_{\pm 0.10}$ & $0.033_{\pm 0.06}$ & $0.089_{\pm 0.15}$ \\
\midrule
CodeLLaMA & latest & 0 & $16.84_{\pm 17.5}$ & $0.052_{\pm 0.09}$ & $0.063_{\pm 0.12}$ & $0.191_{\pm 0.19}$ \\
\midrule
Mistral & latest & 0 & $16.41_{\pm 16.3}$ & $0.109_{\pm 0.11}$ & $0.113_{\pm 0.14}$ & $0.285_{\pm 0.17}$ \\
\midrule
Gemma & 2b & 0 & $10.09_{\pm 10.6}$ & $0.090_{\pm 0.10}$ & $0.038_{\pm 0.07}$ & $0.219_{\pm 0.14}$ \\
\midrule
Starling-LM & latest & 0 & $15.88_{\pm 15.8}$ & $0.093_{\pm 0.10}$ & $0.062_{\pm 0.09}$ & $0.242_{\pm 0.16}$ \\
\midrule
\textbf{Average} 
&  
& $0.04$ 
& $12.37_{\pm 13.83}$ 
& $0.107_{\pm 0.119}$ 
& $0.077_{\pm 0.102}$ 
& $0.213_{\pm 0.154}$ \\
\bottomrule
\end{tabular}
}
\end{table}
\begin{table}[t]
\centering
\caption{SANY error taxonomy across four prompting strategies. Panel~A: distribution of error types among failed runs. Panel~B: location of the first error in the generated file.}
\label{tab:error-taxonomy}

\resizebox{0.32\textwidth}{!}{
\begin{tabular*}{\columnwidth}{@{\extracolsep{\fill}} l c S[table-format=3.0] S[table-format=2.1]@{}}
\multicolumn{4}{c}{\textbf{Panel A: Error Type Distribution}} \\
\midrule
\textbf{Error Type} & \textbf{Method} & \textbf{n} & \textbf{\%} \\
\midrule
Parse: Bad Module Body    & FS   & 228 & 47.8 \\
                         & HC  & 264 & 46.3 \\
                         & Prog. & 243 & 49.8 \\
                         & FIM  & 290 & 52.0 \\
\midrule
Parse: Other              & FS   & 72  & 15.1 \\
                         & HC  & 227 & 39.8 \\
                         & Prog. & 70  & 14.3 \\
                         & FIM  & 191 & 34.2 \\
\midrule
Parse: Unexpected Token   & FS   & 82  & 17.2 \\
                         & HC  & 29  & 5.1  \\
                         & Prog. & 65  & 13.3 \\
                         & FIM  & 17  & 3.0  \\
\bottomrule
\end{tabular*}
}
\vspace{6pt}

\resizebox{0.32\textwidth}{!}{
\begin{tabular*}{\columnwidth}{@{\extracolsep{\fill}}l c S[table-format=3.0] S[table-format=2.1]@{}}
\multicolumn{4}{c}{\textbf{Panel B: First Error Location}} \\
\midrule
\textbf{Location} & \textbf{Method} & \textbf{n} & \textbf{\%} \\
\midrule
Early (first third) & FS   & 240 & 50.3 \\
                    & HC  & 63  & 11.1 \\
                    & Prog. & 220 & 45.1 \\
                    & FIM  & 81  & 14.5 \\
\midrule
Mid (middle third)  & FS   & 196 & 41.1 \\
                    & HC  & 261 & 45.8 \\
                    & Prog. & 209 & 42.8 \\
                    & FIM  & 310 & 55.6 \\
\midrule
Late (final third)  & FS   & 41  & 8.6  \\
                    & HC  & 246 & 43.2 \\
                    & Prog. & 59  & 12.1 \\
                    & FIM  & 167 & 29.9 \\
\bottomrule
\end{tabular*}}
\end{table}

\begin{table}[t]
\centering
\caption{Per-model textual similarity metrics for Half-Completion (HC) prompting (mean$_{\pm}$SD across 26 specifications). Includes BLEU, ROUGE, edit distance, exact match and line accuracy.}
\label{tab:half-completion-summary}
\resizebox{0.48\textwidth}{!}{
\begin{tabular}{@{}l|p{1.75cm}|c|c|c|c|c@{}}
\toprule
\textbf{Family}&   \textbf{Model}&   \textbf{E}&   \textbf{Acc (\%)}&   \textbf{Similarity}&   \textbf{BLEU}&   \textbf{ROUGE-L} \\
\midrule
&   r1:70b&   0&   $14.69_{\pm11.6}$&   $0.154_{\pm0.21}$&   $0.107_{\pm0.17}$&   $0.278_{\pm0.21}$ \\
&   r1:32b&   0&   $15.43_{\pm13.2}$&   $0.175_{\pm0.17}$&   $0.120_{\pm0.16}$&   $0.319_{\pm0.19}$ \\
&   r1:14b&   0&   $11.67_{\pm8.7}$&   $0.083_{\pm0.06}$&   $0.050_{\pm0.07}$&   $0.227_{\pm0.14}$ \\
DeepSeek&   r1:8b&   0&   $4.42_{\pm6.1}$&   $0.008_{\pm0.02}$&   $0.002_{\pm0.01}$&   $0.030_{\pm0.08}$ \\
&   r1:7b&   0&   $12.62_{\pm11.3}$&   $0.081_{\pm0.08}$&   $0.056_{\pm0.07}$&   $0.215_{\pm0.14}$ \\
&   r1:1.5b&   0&   $6.05_{\pm8.0}$&   $0.037_{\pm0.05}$&   $0.020_{\pm0.04}$&   $0.086_{\pm0.13}$ \\
&   coder:6.7b&   0&   $10.48_{\pm14.2}$&   $0.062_{\pm0.10}$&   $0.032_{\pm0.06}$&   $0.163_{\pm0.15}$ \\
&   coder:latest&   0&   $10.80_{\pm11.7}$&   $0.044_{\pm0.04}$&   $0.014_{\pm0.03}$&   $0.109_{\pm0.10}$ \\
\midrule
&   3.3:70b&   0&   $17.67_{\pm13.9}$&   $0.181_{\pm0.17}$&   $0.158_{\pm0.18}$&   $0.366_{\pm0.20}$ \\
&   3.3:latest&   0&   $17.51_{\pm13.5}$&   $0.166_{\pm0.17}$&   $0.161_{\pm0.18}$&   $0.361_{\pm0.20}$ \\
LLaMA&   3.1:8b&   0&   $9.05_{\pm10.7}$&   $0.106_{\pm0.11}$&   $0.083_{\pm0.11}$&   $0.232_{\pm0.16}$ \\
&   3:latest&   0&   $12.07_{\pm9.9}$&   $0.115_{\pm0.12}$&   $0.064_{\pm0.09}$&   $0.265_{\pm0.16}$ \\
&   3.2:1b&   0&   $6.72_{\pm8.7}$&   $0.053_{\pm0.07}$&   $0.043_{\pm0.08}$&   $0.140_{\pm0.14}$ \\
\midrule
&   3:235b&   0&   $5.18_{\pm10.3}$&   $0.061_{\pm0.14}$&   $0.014_{\pm0.05}$&   $0.093_{\pm0.19}$ \\
Qwen&   3:30b&   0&   $3.58_{\pm4.8}$&   $0.041_{\pm0.15}$&   $0.023_{\pm0.08}$&   $0.058_{\pm0.21}$ \\
&   3:4b&   0&   $8.24_{\pm9.2}$&   $0.127_{\pm0.15}$&   $0.067_{\pm0.15}$&   $0.217_{\pm0.21}$ \\
\midrule
QwQ&   latest&   0&   $6.06_{\pm9.2}$&   $0.037_{\pm0.05}$&   $0.018_{\pm0.02}$&   $0.066_{\pm0.04}$ \\
\midrule
Phi&   4-mini:latest&   0&   $1.54_{\pm4.2}$&   $0.015_{\pm0.03}$&   $0.003_{\pm0.01}$&   $0.034_{\pm0.06}$ \\
\midrule
Granite&   3.3:latest&   0&   $15.30_{\pm12.6}$&   $0.071_{\pm0.07}$&   $0.055_{\pm0.10}$&   $0.192_{\pm0.15}$ \\
\midrule
GPT-OSS&   120b&   0&   $18.39_{\pm14.1}$&   $0.195_{\pm0.17}$&   $0.151_{\pm0.18}$&   $0.362_{\pm0.22}$ \\
&   20b&   0&   $11.12_{\pm12.9}$&   $0.098_{\pm0.13}$&   $0.052_{\pm0.10}$&   $0.197_{\pm0.22}$ \\
\midrule
CodeLLaMA&   latest&   0&   $8.84_{\pm11.2}$&   $0.057_{\pm0.11}$&   $0.038_{\pm0.09}$&   $0.121_{\pm0.16}$ \\
\midrule
Mistral&   latest&   0&   $14.15_{\pm12.3}$&   $0.068_{\pm0.06}$&   $0.084_{\pm0.11}$&   $0.213_{\pm0.17}$ \\
\midrule
Gemma&   2b&   0&   $15.23_{\pm11.5}$&   $0.073_{\pm0.06}$&   $0.067_{\pm0.09}$&   $0.194_{\pm0.11}$ \\
\midrule
Starling-LM&   latest&   0&   $15.73_{\pm12.3}$&   $0.083_{\pm0.09}$&   $0.070_{\pm0.12}$&   $0.188_{\pm0.13}$ \\
\midrule
\textbf{Total}&   --&   0&   $10.90_{\pm10.6}$&   $0.088_{\pm0.10}$&   $0.062_{\pm0.09}$&   $0.189_{\pm0.15}$ \\
\bottomrule
\end{tabular}
}
\end{table}

\subsection{RQ3: Model Impact}

\UlPara{Model Family Analysis} The per-model results in 
Table~\ref{tab:all-methods-results} reveal that model family predicts 
\textit{which} strategy a model excels at more than it predicts overall 
quality. The \textit{Qwen3} family dominates syntax generation under 
full-generation strategies but achieves no TLC passes under FS despite 
leading the SANY rankings. High syntactic fluency does not translate to 
semantic correctness without the structured guidance of Prog. The 
\textit{DeepSeek R1} series presents the most striking result: the 8B 
parameter model achieves the best TLC result in the entire study (14/26 
under Prog.) while the 70B variant fails to produce a single SANY pass under 
FS. This inversion of scale is not noise: it appears consistently across 
metrics and strategies (see Model Size below). The R1 training regime produces 
reasoning capabilities that transfer differently at different scales. The 
\textit{LLaMA~3.3} family performs reliably under Prog.\ for syntax (18/26 
for both 70b and latest) but, like Qwen, achieves no TLC passes.

\UlPara{Model Size} Larger models do not consistently outperform smaller 
ones, a finding that holds within three independent model families. Within 
DeepSeek R1, the 8B model achieves 14 TLC passes under Prog.\ while the 70B 
model achieves none. The 8B model also leads on SANY under FS (22/26 vs.\ 
0/26). Within Qwen3, the 30B variant leads FS SANY (23/26) over the 235B 
model (22/26). Within LLaMA, the 1B model (3.2:1b) achieves the highest 
Prog.\ SANY of any LLaMA variant (12/26), ahead of LLaMA 3.1:8b (3/26) and 
LLaMA 3:latest (2/26). For \tla, a scarce-corpora formal language, 
reasoning alignment and instruction-following discipline matter more than raw 
parameter count.

\UlPara{Code Specialization} Models designed or fine-tuned for code 
generation do not outperform general-purpose models on \tla. CodeLLaMA 
achieves 0--5 SANY passes per strategy, DeepSeek-Coder peaks at 5/26 (HC and 
FIM), and Granite reaches only 5/26 (HC). General-purpose models like 
Qwen3:30b (23/26 FS) and LLaMA 3.3:70b (18/26 Prog.) far exceed these code 
specialists. Models heavily fine-tuned on Python, C, and Java learn strong 
priors about those languages' syntax that interfere with the 
stricter and more distinctive grammar of \tla.

\UlPara{Per-Specification Difficulty} Table~\ref{tab:spec-summary} shows 
that specification difficulty varies from 0\% to 100\% SANY pass rates and 
that the difficulty profile is strategy-specific. Simple state-machine 
specifications (\eg \texttt{YoYoNoPruning}) achieve 100\% SANY under both 
completion strategies, while complex distributed protocols (\eg 
\texttt{HDiskSynod}, \texttt{BinarySearch}) score 0\%. Under Prog.\ with TLC, 
\texttt{MultiPaxos} and \texttt{VoucherCancel} achieve the highest semantic 
pass rates (16\% each). \texttt{MCDieHarder} achieves 0\% TLC across every 
strategy. 
\begin{tcolorbox}[title=\textbf{RQ3 Summary}]
Standard assumptions about scale and specialization do not hold for 
\tla. Smaller reasoning models (DeepSeek r1:8b, LLaMA 3.2:1b) outperform larger 
siblings. Code-specialist models consistently underperform general-purpose 
ones due to negative transfer from mainstream language training. Model family 
determines strategy fit better than overall capability. 
\end{tcolorbox}

\begin{table}[t]
\centering
\caption{Specification-level SANY and TLC pass rates averaged across 25 open-weight models for each of the 26 test specifications. Per-specification similarity metrics and detailed breakdowns are provided in supplementary materials.}
\label{tab:spec-summary}

\resizebox{0.48\textwidth}{!}{
\begin{tabular}{@{}l|cc|cc|cc|cc@{}}
\toprule
 & \multicolumn{2}{c}{\textbf{FS}} & \multicolumn{2}{c}{\textbf{HC}} & \multicolumn{2}{c}{\textbf{Prog.}} & \multicolumn{2}{c}{\textbf{FIM}} \\
\cmidrule(lr){2-3} \cmidrule(lr){4-5} \cmidrule(lr){6-7} \cmidrule(lr){8-9}
\textbf{Specification} & \textbf{S} & \textbf{T} & \textbf{S} & \textbf{T} & \textbf{S} & \textbf{T} & \textbf{S} & \textbf{T} \\
\midrule
Alternate & 44.0& 0.0& 4.0& 0.0& 28.0& 4.0& 0.0& 0.0\\
BinarySearch & 12.0& 0.0& 0.0& 0.0& 24.0& 8.0& 0.0& 0.0\\
BufferedRandom\\AccessFile & 16.0& 0.0& 0.0& 0.0& 28.0& 12.0& 0.0& 0.0\\
ClientCentric & 24.0& 0.0& 16.0& 0.0& 12.0& 8.0& 0.0& 0.0\\
Consensus & 44.0& 0.0& 4.0& 0.0& 36.0& 12.0& 8.0& 0.0\\
Environment\\Controller & 24.0& 0.0& 4.0& 0.0& 24.0& 16.0& 4.0& 0.0\\
EWD687a\_anim & 36.0& 0.0& 0.0& 0.0& 28.0& 12.0& 4.0& 0.0\\
EWD998ChanID\\\_export & 20.0& 0.0& 0.0& 0.0& 16.0& 8.0& 0.0& 0.0\\
Hanoi & 4.0& 0.0& 20.0& 0.0& 16.0& 12.0& 4.0& 0.0\\
HDiskSynod & 32.0& 0.0& 0.0& 0.0& 20.0& 4.0& 48.0& 0.0\\
KVsnap & 28.0& 0.0& 16.0& 0.0& 24.0& 12.0& 0.0& 0.0\\
Majority & 20.0& 0.0& 8.0& 0.0& 16.0& 4.0& 8.0& 0.0\\
MajorityProof & 32.0& 0.0& 32.0& 0.0& 28.0& 4.0& 12.0& 0.0\\
MCConsensus & 36.0& 0.0& 4.0& 0.0& 40.0& 8.0& 8.0& 0.0\\
MCDieHarder & 28.0& 0.0& 4.0& 0.0& 16.0& 0.0& 24.0& 0.0\\
MCEWD687a & 24.0& 0.0& 24.0& 0.0& 32.0& 4.0& 36.0& 0.0\\
MultiPaxos & 24.0& 0.0& 8.0& 0.0& 32.0& 16.0& 0.0& 0.0\\
MultiPaxos\_MC & 36.0& 0.0& 36.0& 0.0& 36.0& 4.0& 8.0& 0.0\\
nbacc\_ray97 & 32.0& 0.0& 0.0& 0.0& 20.0& 12.0& 36.0& 0.0\\
Relation & 28.0& 0.0& 4.0& 0.0& 20.0& 4.0& 8.0& 0.0\\
SimpleRegular & 24.0& 0.0& 8.0& 0.0& 32.0& 12.0& 44.0& 0.0\\
TLAPS & 20.0& 0.0& 12.0& 0.0& 24.0& 4.0& 4.0& 0.0\\
TwoPhase & 24.0& 0.0& 8.0& 0.0& 20.0& 8.0& 12.0& 0.0\\
VoucherCancel & 24.0& 0.0& 4.0& 0.0& 28.0& 16.0& 0.0& 0.0\\
VoucherTransfer & 32.0& 0.0& 4.0& 0.0& 24.0& 4.0& 0.0& 0.0\\
YoYoNoPruning & 24.0& 0.0& 100.0& 0.0& 24.0& 16.0& 100.0& 0.0\\
\midrule

\textbf{Average} 
& \textbf{26.6} 
& \textbf{0.0} 
& \textbf{12.2} 
& \textbf{0.0} 
& \textbf{24.9} 
& \textbf{8.6} 
& \textbf{14.2} 
& \textbf{0.0} \\
\bottomrule
\end{tabular}
}
\end{table}

\begin{table}[t]
\centering
\caption{Hallucination taxonomy across prompting strategies (RQ4). Integer values = file counts; $\checkmark$ = observed but count not isolated; 0 = not observed; n/a = not applicable. $^\dagger$FIM applies \texttt{clean\_llm\_response()} to strip markdown before validation. $^\ddagger$File scan (contains $\geq$1 semicolon). $^\S$SANY first-error count (FS: 18/477; HC: 11/570; Prog. : 25/488; FIM: 10/558). $^\P$File scan shows higher rates (FS: 49/650 = 7.5\%; HC: 23/650 = 3.5\%; Prog. : 82/650 = 12.6\%; FIM: 22/650 = 3.4\%).}
\label{tab:hallucination-taxonomy}

\resizebox{\columnwidth}{!}{
\begin{tabular}{@{}llp{1.5cm}cccc@{}}
\toprule
\textbf{Pattern} & \textbf{Source} & \textbf{Correct TLA+} & \textbf{FS} & \textbf{HC} & \textbf{Prog} & \textbf{FIM} \\
\midrule

\multicolumn{7}{l}{\textit{Unicode Operator Substitution (Lexical Error)}} \\
\midrule
$\lor$ (U+2228) & \LaTeX/Math & \texttt{\textbackslash/} & $\checkmark$ & $\checkmark$ & $\checkmark$ & $\checkmark$ \\
$\land$ (U+2227) & \LaTeX/Math & \texttt{/\textbackslash} & $\checkmark$ & $\checkmark$ & $\checkmark$ & $\checkmark$ \\
$\in$ (U+2208) & \LaTeX/Math & \texttt{\textbackslash in} & $\checkmark$ & $\checkmark$ & $\checkmark$ & $\checkmark$ \\
$\to$ (U+2192) & \LaTeX/Math & \texttt{->} & $\checkmark$ & $\checkmark$ & $\checkmark$ & $\checkmark$ \\

\midrule
\multicolumn{7}{l}{\textit{Cross-Language Syntax Injection}} \\
\midrule
Semicolons (\texttt{;}) & C/Java/Py & \textit{invalid} & 18$^\S$/49$^\P$ & 11$^\S$/23$^\P$ & 25$^\S$/82$^\P$ & 10$^\S$/22$^\P$ \\
Backticks (\texttt{`}) & Markdown & \textit{invalid} & 28 & 10 & 21 & 13 \\
\texttt{END} keyword & Pascal/SQL & \texttt{====} & 0 & 0 & $\checkmark$ & 0 \\

\midrule
\multicolumn{7}{l}{\textit{Reasoning/Formatting Leakage}} \\
\midrule
\texttt{<think>} blocks & CoT reasoning & \textit{invalid} & 15 & 30 & n/a & 28 \\
Markdown fences & Output format & \textit{invalid} & 0 & 0 & $\checkmark$ & 0$^\dagger$ \\
NL prose in body & Task confusion & \textit{invalid} & 0 & 0 & $\checkmark$ & 0 \\

\midrule
\multicolumn{7}{l}{\textit{Structural Errors}} \\
\midrule
Missing \texttt{====} & Incomplete gen & \texttt{====} & $\checkmark$ & $\checkmark$ & 243+ & $\checkmark$ \\
Missing MODULE & No preamble & \texttt{- MODULE X -} & 0 & 0 & $\checkmark$ & 0 \\
Duplicate MODULE & Repeated start & \textit{single header} & 0 & 0 & $\checkmark$ & 0 \\

\bottomrule
\end{tabular}
}

\end{table}

\subsection{RQ4: Failure Modes}

Beyond the error type distribution in RQ2, we identify five categories of 
hallucination that recur across models and strategies (Table~\ref{tab:hallucination-taxonomy}).

\UlPara{Unicode Operator Substitution} LLMs frequently replace the ASCII 
operators that \tla requires with their Unicode or \LaTeX\ equivalents. 
Common substitutions include $\lor$ (U+2228) for \texttt{\textbackslash/}, 
$\land$ (U+2227) for \texttt{/\textbackslash}, $\in$ (U+2208) for 
\texttt{\textbackslash in}, and $\to$ (U+2192) for \texttt{->}. These 
substitutions appear across all four strategies and all model families. SANY 
rejects all Unicode operators, which makes this a consistent source of lexical 
errors.

\UlPara{Cross-Language Syntax Injection} Models inject syntax from 
mainstream languages into \tla output. \textit{Semicolons} are the most 
common example: Prog.\ shows the highest contamination rate at 12.6\% of 
files (82/650) followed by FS at 7.5\% (49/650), while HC and FIM are lower 
at approximately 3.5\% each. 
QwQ is the 
worst offender with 35\% semicolon contamination in FS and 27\% in Prog. 
Granite and Mistral each reach 31\% in Prog. \textit{Backtick injection} from 
Markdown code fences is most common in FS (28 files), with DeepSeek r1:7b 
particularly prone (12 out of 26 files). Prog.\ also shows isolated instances 
of \texttt{END} keyword injection from Pascal/SQL syntax.

\UlPara{Reasoning and Formatting Leakage} Models with chain-of-thought 
capabilities leak their internal reasoning into the generated specification. 
QwQ is the worst offender: it injects \texttt{<think>} blocks into 100\% of 
its HC and FIM outputs (26/26 files each) and 46.2\% of FS outputs 
DeepSeek R1 models (r1:14b and r1:32b) 
and Qwen3:235b show lower contamination rates (3.8--7.7\%). The overall 
contamination rates are 2.3\% for FS, 4.6\% for HC, and 4.3\% for FIM. 
Prog.\ is not affected because its multi-step pipeline structure handles 
reasoning differently. Beyond thinking blocks, Prog.\ outputs occasionally 
contain Markdown fences and natural language prose within the \tla 
module body.
\begin{tcolorbox}[title=\textbf{RQ4 Summary}]
Five systematic hallucination categories recur across all models: 
(1)~Unicode operator substitution; (2)~cross-language syntax injection 
(semicolons in 12.6\% of files, backticks in 5.9\%); (3)~reasoning token 
leakage (\texttt{<think>} blocks in QwQ); (4)~generation length 
miscalibration (0.00--9.84$\times$ ground truth); and (5)~structural errors 
(243+ missing \texttt{====} terminators). Each category is traceable to a 
specific training data bias and addressable by better datasets, 
grammar-specific decoding/post-processing.
\end{tcolorbox}
\UlPara{Generation Length Miscalibration} Models exhibit poor calibration 
between output length and the expected ground-truth length 

On average, HC outputs are 3.46$\times$ the 
length of the ground-truth target segment and FIM outputs are 2.24$\times$. 
The extremes are stark: DeepSeek coder:latest generates 9.15$\times$ the 
expected HC length, LLaMA 3.2:1b generates 9.84$\times$, and QwQ generates 
7.71$\times$ for HC and 7.75$\times$ for FIM. At the opposite extreme, 
Qwen3:30b produces near-empty output for HC (0.00$\times$) and Qwen3:235b 
produces only 0.09$\times$ the expected length. Only a few models achieve 
near-perfect calibration: LLaMA 3.3:70b (1.16$\times$ HC and 1.03$\times$ 
FIM) and DeepSeek r1:7b (0.97$\times$ HC).

\UlPara{Structural Errors} Prog.\ produces distinctive structural 
hallucinations due to its multi-step generation process. The most common is 
the missing \texttt{====} terminator: over 243 Prog.-generated files lack the 
four-equals-sign module terminator that \tla requires. Models also 
occasionally produce duplicate \texttt{MODULE} headers or omit the 
\texttt{MODULE} header entirely when the focus is on body content from a 
particular step. These structural errors are largely absent from the other strategies where generation occurs in one pass.

\section{Discussion}\label{sec:discussion}

This section discusses the results and practical implications for \tla LLM-based synthesis.

\UlPara{The Syntax-Semantics Gap Is Structural, Not Incidental}
The wide gap between syntactic validity (up to 26.6\% SANY) and semantic correctness (8.6\% TLC)  persists across all four strategies and all model scales.
It reflects a structural property: LLMs acquire surface-level token patterns of \tla syntax without internalizing the modal and set-theoretic semantics those tokens encode.
A \tla module that parses is only a skeleton.
Hahn~et~al.\ show that LLM performance degrades as target formalism complexity increases from regular expressions through first-order logic to LTL~\cite{hahn_formal_nodate}; \tla, combining TLA with set theory, sits at the harder end of that spectrum.
Ferrari and Spoletini frame LLMs and formal methods as mutually reinforcing: LLMs accelerate specification drafting while formal tools provide correctness guarantees~\cite{ferrari_formal_2025}. Our results illustrate why both directions are needed.
The Qwen3 family makes the gap most concrete: Qwen3:30b achieves the highest FS SANY rate (23/26) yet produces zero TLC passes under the same strategy.
The results establish that closing the syntax-semantics gap is the primary challenge for future \tla work. 
Fine-tuning on the 205 \tla Foundation specifications could improve semantic understanding, and iterative refinement with TLC feedback could help toward valid synthesis.

\UlPara{Progressive Prompting as Cognitive Scaffolding}
Progressive prompting is the only strategy that achieves TLC passes (8.6\%) and its SANY rate (24.9\%) is competitive with FS (26.6\%), making it the dominant choice overall.
Its advantage lies in decomposing the generation task into sequential concerns: module structure, variable declarations, initial predicate, next-state relation, temporal property.
This matches the insight behind chain-of-thought prompting, where externalising intermediate reasoning steps improves performance on tasks requiring multi-hop inference~\cite{wei2022chain}.
Correct \tla synthesis must simultaneously satisfy module syntax, type consistency, behavioral semantics and temporal logic.
Kogler~et~al.\ independently found that iterative LLM prompting with post-processing achieves syntactic validity in formal railway specifications~\cite{kogler_reliable_2024}, and the Specula system's architecture confirms that multi-component pipelines substantially outperform single-pass generation~\cite{specula2025}.
The cost of progressive prompting is a new class of structural errors (duplicate headers, missing terminators) that do not arise in single-pass strategies. This observation suggests that future systems should pair progressive decomposition with a lightweight global consistency check.
The baseline 8.6\% TLC success under progressive prompting sets a foundation that future work with fine-tuned models and improved datasets could substantially exceed through  progressive decomposition and semantic feedback.

\UlPara{Model Scale and the Scarce-corpora Formal Language Problem.}
The inversion of scale, where DeepSeek r1:8b outperforms r1:70b and Qwen3:30b matches Qwen3:235b, recurs across families and strategies and is not noise.
\tla is a scarce-corpora formal language: its entire public corpus amounts to hundreds of modules, dwarfed by billions of tokens of C, Python and Java.
Larger models have proportionally more capacity devoted to high-resource languages, which manifests as stronger prior pulls toward mainstream syntax.
Smaller reasoning-distilled models in the DeepSeek R1 series were trained with reinforcement learning on step-by-step symbolic reasoning traces~\cite{Guo_2025}; at 8B parameters the model reasons over \tla structure without the noise of the larger conflicting training pool. Mirzadeh~et~al.\ show that LLMs replicate reasoning steps from training data rather than performing genuine logical reasoning~\cite{mirzadeh_gsm-symbolic_2025}. A model whose reasoning depends on training distribution will be most affected when that distribution is sparse, as it is for \tla.
For practitioners, this baseline guidance means that the largest available model is not necessarily the right choice: smaller, reasoning-oriented models may be more reliable and substantially cheaper.

\UlPara{Negative Transfer from Code Specialization}
Code-specialist models (CodeLLaMA, DeepSeek-Coder, Granite) consistently underperform general-purpose counterparts across all strategies.
The mechanism is negative transfer: fine-tuning on C, Java and Python trains the model to associate token sequences (semicolons, curly braces, \texttt{return}) with correct code, and these associations are activated when generating \tla. 
This directly accounts for two of our five hallucination categories: semicolon injection and backtick fence insertion. Beg~et~al.\ identify the mismatch between target formal languages and pre-training distributions as a principal challenge for LLM-based formal requirements~\cite{beg_leveraging_2025}.
When selecting a model for \tla generation, general-purpose reasoning-oriented models retain the linguistic flexibility needed for \tla's unusual operator vocabulary and set-theoretic structure.

\UlPara{Hallucinations Are Systematic and Mitigable}
All five hallucination categories are traceable to specific training distributions rather than to the \tla generation task itself.
Unicode operator substitution arises because documentation renders operators in mathematical Unicode ($\in$, $\land$) while SANY requires ASCII (\texttt{\textbackslash in}, \texttt{/\textbackslash}).
Semicolon and backtick injection arise from C-family and Markdown-dominated pre-training corpora.
Reasoning token leakage is a side effect of chain-of-thought fine-tuning. Generation length miscalibration (3.46$\times$ for HC, 2.24$\times$ for FIM, with outliers at 9.84$\times$) directly explains the near-zero textual similarity scores. Phantom operator invention parallels the package hallucination phenomenon~\cite{spracklen2025packagehallucinations}: models generate identifiers statistically consistent with context that do not exist in the target environment.
The systematic and predictable nature of these errors is itself an opportunity: grammar-constrained decoding~\cite{helwer2025} can enforce \tla syntax at the token level, deterministic post-processing (Unicode normalization, semicolon stripping, fence removal) can eliminate a large fraction of SANY failures before the parser is invoked, and retrieval-augmented generation over verified \tla specifications can ground operator usage in concrete examples. 

\UlPara{Implications for Practice}
Progressive prompting achieves semantically valid specifications.
Reasoning-oriented is favored over general-purpose models and code specialists. 
Deterministic post-processing (with augmented retrieval) before SANY could help with \tla synthesis, \eg Unicode normalization, fence removal, and reasoning token stripping to eliminate the majority of preventable syntax failures at zero cost.
SANY pass rate can be only a base, as semantic validity requires model checking.

\UlPara{Limitations}
We run each model-specification pair once per strategy, so stochastic variation is not fully captured; we address this through scale (650 runs per strategy, 2,600 core runs total) with temperature and sampling parameters held constant across all models. Our 26 test specifications cover distributed protocols, concurrency problems, and algorithmic puzzles from the \tla community repository, but may not represent the full range of industrial use cases, as proprietary specifications from organizations such as Amazon and Microsoft are not publicly available. This study focuses on \tla, and the results may not transfer directly to other formal languages. Finally, the natural language inputs vary in comment density across specifications, and TLC validation is bounded to finite-state instances, so errors that appear only with larger parameter values remain undetected.

\section{Conclusion} \label{sec:conclusion}

We present the first systematic evaluation of LLM-based \tla specification synthesis from natural language. Across 30 models, eight families, 26 test specifications, and more than 2,700 runs, results show a clear syntax-semantics gap. 
LLMs reach 26.6\% SANY validity, yet only 8.6\% pass TLC, and only under progressive prompting. DeepSeek r1:8b outperforms its 70B counterpart. Code-specialized models also trail general-purpose models. 
Through our experiments, we identify five recurring hallucination categories: Unicode operator substitution, cross-language syntax injection, reasoning leakage, length miscalibration, and structural errors. These results show that current LLMs do not yet produce reliable \tla specifications rigorous review by a human.
Future work should target three challenges in this area: (1)\ iterative repair/feedback loop with SANY and TLC to address the syntax-semantics gap; (2)\ multi-step prompting with global consistency checks to mitigate lexical and structural errors in progressive generation; and (3)\ fine-tuning on \tla data with grammar-constrained decoding to reduce negative transfer and systematic hallucinations. 
\balance
\bibliographystyle{apalike}
{\small
\bibliography{custom}}

\end{document}